\pdfoutput=1

\documentclass[11pt]{article}

\usepackage{latex/acl}

\usepackage{times}
\usepackage{latexsym}

\usepackage[T1]{fontenc}

\usepackage[utf8]{inputenc}

\usepackage{microtype}

\usepackage{inconsolata}

\usepackage[nolist]{acronym}
\usepackage{graphicx}
\usepackage{multirow}
\usepackage{subcaption}
\usepackage{adjustbox,booktabs}
\usepackage{dblfloatfix} 
\usepackage{xcolor}
\usepackage{booktabs}
\usepackage{enumitem}
\usepackage{color}
\usepackage{pifont}
\usepackage{array}
\usepackage[most]{tcolorbox}

\newcommand{\cmark}{\textcolor{green!50!black}{\ding{51}}} 
\newcommand{\xmark}{\textcolor{red}{\ding{55}}} 

\setlist[itemize]{itemsep=0em, topsep=0em}

\title{NLP-KG: A System for Exploratory Search of Scientific Literature in Natural Language Processing}

\author{Tim Schopf \and Florian Matthes \\
         Technical University of Munich, Department of Computer Science, Germany \\
         \texttt{\{tim.schopf,matthes\}@tum.de}}

\begin{acronym}
\acro{nlp}[NLP]{natural language processing}
\acro{rag}[RAG]{Retrieval Augmented Generation}
\acro{lm}[LM]{Language Model}
\acro{llm}[LLM]{Large Language Model}
\acroplural{llm}[LLMs]{Large Language Models}
\acro{plm}[PLM]{Pretrained Language Model}
\acroplural{plm}[PLMs]{Pretrained Language Models}
\acro{nlp-kg}[NLP-KG]{Natural Language Processing Knowledge Graph}
\acro{kg}[KG]{knowledge graph}
\acroplural{kg}[KGs]{knowledge graphs}
\acro{p}[P]{Precision}
\acro{r}[R]{Recall}
\acro{mag}[MAG]{Microsoft Academic Graph}
\acro{fos}[FoS]{Field of Study}
\acroplural{fos}[FoS]{Fields of Study}
\acro{mag}[MAG]{Microsoft Academic Graph}
\acro{cso}[CSO]{Computer Science Ontology}
\acro{ai-kg}[AI-KG]{Artificial Intelligence Knowledge Graph}
\acro{orkg}[ORKG]{Open Research Knowledge Graph}
\acro{plmarker}[PL-Marker]{Packed Levitated Marker}
\acro{mape}[MAPE]{Mean Absolute Percentage Error}
\end{acronym}

\begin{document}
\maketitle
\begin{abstract}

Scientific literature searches are often exploratory, whereby users are not yet familiar with a particular field or concept but are interested in learning more about it. However, existing systems for scientific literature search are typically tailored to keyword-based lookup searches, limiting the possibilities for exploration. We propose NLP-KG, a feature-rich system designed to support the exploration of research literature in unfamiliar natural language processing (NLP) fields. In addition to a semantic search, NLP-KG allows users to easily find survey papers that provide a quick introduction to a field of interest. Further, a Fields of Study hierarchy graph enables users to familiarize themselves with a field and its related areas. Finally, a chat interface allows users to ask questions about unfamiliar concepts or specific articles in NLP and obtain answers grounded in knowledge retrieved from scientific publications. Our system provides users with comprehensive exploration possibilities, supporting them in investigating the relationships between different fields, understanding unfamiliar concepts in NLP, and finding relevant research literature. Demo, video, and code are available at: \href{https://github.com/NLP-Knowledge-Graph/NLP-KG-WebApp}{https://github.com/NLP-Knowledge-Graph/NLP-KG-WebApp}.




\end{abstract}

\section{Introduction}

\begin{figure}[ht!]
    \centering
    \resizebox{\columnwidth}{!}{%
    \includegraphics{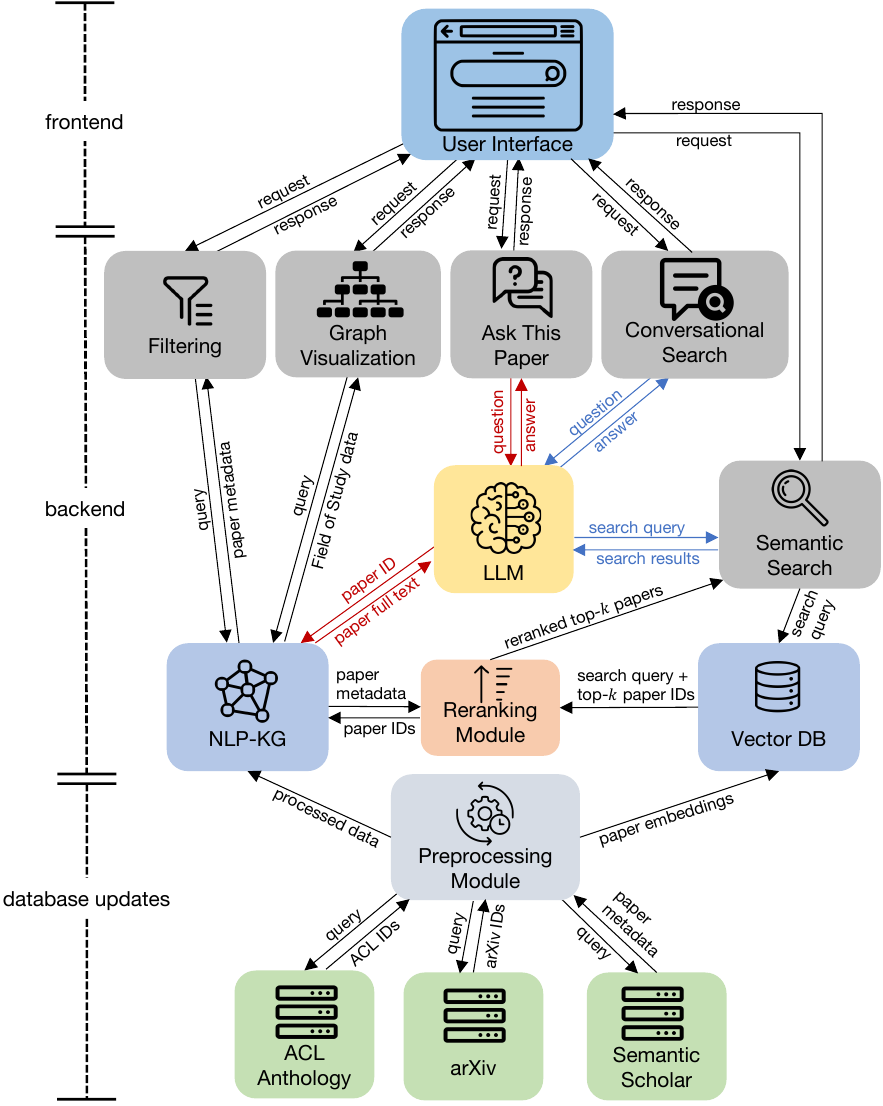}}
    \caption{The architecture of our system. The direction of an arrow represents the direction of data flow. The red arrows show how the autoregressive Large Language Model (LLM) routes the data for the \textit{Ask This Paper} feature, while the blue arrows show how the LLM routes the data for the \textit{Conversational Search} feature. The preprocessing module regularly fetches new publications and processes them to update the knowledge graph and the vector database.}
    \label{fig:system-architecture}
\end{figure}

The body of \ac{nlp} literature has experienced remarkable growth in recent years, with articles on various topics and applications being published in an increasing number of journals and conferences \cite{schopf-etal-2023-exploring}. To browse and search the increasing amount of \ac{nlp}-related literature, researchers may use systems such as Google Scholar\footnote{\href{https://scholar.google.com}{https://scholar.google.com}} or Semantic Scholar \cite{kinney2023semantic}. Both systems cover a wide variety of academic disciplines. Although this has advantages, the lack of focus on \ac{nlp} literature also has disadvantages, e.g., the potential to retrieve lots of search results containing many irrelevant papers \cite{mohammad-2020-nlp-scholar}. For example, when interested in \ac{nlp} literature on \textit{emotion} or \textit{privacy}, searching for it on Google Scholar is less efficient than searching for it on a platform dedicated to \ac{nlp} literature. Further, scholarly literature searches are often exploratory, whereby users are not yet familiar with a particular field or concept and are interested in learning more about it \cite{10.1145/3498366.3505818}. However, commonly used search systems are usually optimized for targeted lookup searches, limiting search and exploration to keyword-based searches and citation-based exploration.


% ... 


In this paper, we present a system to support the exploration of \ac{nlp} research literature from unfamiliar fields using a \ac{kg} and state-of-the-art retrieval approaches. Our main contributions comprise the following features:

\begin{itemize}
    \item \textbf{Graph visualization} of hierarchically structured \acp{fos} in \ac{nlp}. \acp{fos} are academic disciplines and concepts, commonly comprised of (but not limited to) tasks or methods \cite{shen-etal-2018-web}. The graph visualization offers researchers new to a field a starting point for their exploration and supports them to familiarize themselves with a field and its related areas.

    \item \textbf{Semantic search} provides a familiar interface to enable keyword-based searches for publications, authors, venues, and \acp{fos} in \ac{nlp}.

    \item \textbf{Conversational search} responds to \ac{nlp}-related user questions in natural language and grounds the answers in knowledge from academic publications using a \ac{rag} pipeline. This feature allows users to ask questions about unfamiliar concepts and fields in \ac{nlp} and provides explanations as well as reference literature for further exploration.

    \item \textbf{Ask this paper} uses an autoregressive \ac{llm} to answer in-depth user questions about specific publications based on their full texts. This can support users to understand papers from unfamiliar fields. 

    \item \textbf{Advanced filters} can filter the search results for specific \acp{fos}, venues, dates, citation counts, or survey papers. Especially filtering by survey papers can support users to quickly get an introduction to their field of interest.
\end{itemize}

Our system is not intended to replace commonly used search engines but to serve as a supplementary tool for dedicated exploratory search of \ac{nlp} research literature. 

\section{Related Work}

Google Scholar, Semantic Scholar \cite{kinney2023semantic}, ArnetMiner \cite{10.1145/1401890.1402008}, \ac{mag} \cite{10.1145/2740908.2742839,10.1162/qss_a_00021}, OpenAlex \cite{priem2022openalex}, and \ac{orkg} \cite{10.1145/3360901.3364435,AuerOelenHarisStockerDSouzaFarfarVogtPrinzWiensJaradeh+2020+516+529} are all systems for search and discovery of academic literature covering a wide range of scholarly domains. 

\citet{weitz-schafer-2012-graphical} focus on citation analyses of \ac{nlp}-related literature. CL Scholar \cite{singh-etal-2018-cl} is a system that can answer binary, statistical, and list-based queries about computational linguistics publications. Additionally, \ac{nlp} Scholar \cite{mohammad-2020-nlp-scholar} provides interactive visualizations of venues, authors, n-grams, and keywords extracted from \ac{nlp}-related publications, while the \ac{nlp} Explorer \cite{10.1007/978-3-030-45442-5_61} provides \ac{fos} tags and temporal statistics to search and explore the field of \ac{nlp}.

\section{NLP-KG}

A well-organized hierarchical structure of \acp{fos} and an accurate mapping between these \acp{fos} and scholarly publications can enable a streamlined and satisfactory exploration experience \cite{shen-etal-2018-web}. Further, semantic relations between scholarly entities can be easily modeled in a graph representation. Therefore, we construct the \ac{nlp-kg} as the core of our system that links \acp{fos}, publications, authors, and venues via semantic relations. In addition, we integrate a \ac{llm} in our retrieval pipeline that can enhance the exploration experience by providing accurate responses to user queries \cite{zhu2024large}. Figure \ref{fig:system-architecture} illustrates how the knowledge graph and the \ac{llm} are integrated into our system. 

\subsection{Fields of Study Hierarchy Construction} \label{sec:fos-construction}

During exploration, users typically navigate from more well-known general concepts to less well-known and more specific concepts. Therefore, we use a semi-automated approach to construct a high-quality, hierarchical, acyclic graph of \ac{fos} in \ac{nlp}. As a starting point, we use a readily available high-level taxonomy of concepts in \ac{nlp} \cite{schopf-etal-2023-exploring}. At the top level, this \ac{nlp} taxonomy includes 12 different concepts covering the wide range of \ac{nlp}, and consequently, additional concepts can be considered as hyponyms thereof. In total, this \ac{nlp} taxonomy already includes 82 different \acp{fos}, to which we subsequently add further \acp{fos} as hyponyms and co-hyponyms. 

\paragraph{Automated Knowledge Extraction}

For automated extraction of \acp{fos} and hierarchical relations, we use a corpus of titles and abstracts of research publications from the ACL Anthology\footnote{\href{https://aclanthology.org}{https://aclanthology.org}} and the cs.CL category of arXiv\footnote{\href{https://arxiv.org}{https://arxiv.org}}. After removing duplicates, the corpus includes a total of 116,053 documents. For entity and relation extraction, we fine-tune \ac{plmarker} models \cite{ye-etal-2022-packed} on a slightly adapted SciERC dataset \cite{luan-etal-2018-multi}. Since we do not distinguish between different entity types in our \ac{fos} hierarchy graph, we process the SciERC dataset to unify all entity types and transform the original named entity recognition task into a more simple entity extraction task. Additionally, we only use the \textit{Hyponym-of} relationship to extract hierarchical relations. Finally, we experiment with BERT \cite{devlin-etal-2019-bert}, SciBERT \cite{beltagy-etal-2019-scibert}, SPECTER2 \cite{singh-etal-2023-scirepeval}, and SciNCL \cite{ostendorff-etal-2022-neighborhood} as base models. 

\begin{table}[!ht]
    \centering
    \resizebox{\columnwidth}{!}{%
   

    \begin{tabular}{lc ccc ccc}
    \toprule
    \multicolumn{2}{l}{\textbf{Task $\rightarrow$}} & \multicolumn{3}{c}{\textit{Entity Extraction}} & \multicolumn{3}{c}{\textit{Relation Extraction}} \\
    
    \cmidrule(lr){3-5}
    \cmidrule(lr){6-8}
    
    \multicolumn{2}{l}{\textbf{Model $\downarrow$}} &      \textbf{P} &      \textbf{R} &    $\textbf{F}_{1}$ &            \textbf{P} &      \textbf{R} &    $\textbf{F}_{1}$  \\
    
    \midrule

    \multicolumn{2}{l}{BERT} &
    68.87 &  66.63 & 67.73 & 70.01 & 68.28 & 69.13 \\
    \multicolumn{2}{l}{SciBERT} & 
    69.91 & \textbf{67.09} & \textbf{68.47} & 71.23 & \textbf{69.63} & \textbf{70.42} \\
    \multicolumn{2}{l}{SPECTER2} & 
    \textbf{69.99} & 66.52 & 68.21 & 69.66 & 68.95 & 69.30 \\
    \multicolumn{2}{l}{SciNCL} & 
    69.59 & 65.39 & 67.42 & \textbf{71.24} & 68.28 & 69.73 \\
  
    \bottomrule
    \end{tabular}
}
\caption{Evaluation results for \ac{plmarker} fine-tuning on the processed SciERC test set using different base models. We report micro (P)recision, (R)ecall, and ${F}_{1}$ scores.}
\label{tab:extraction_results}
\end{table}

The evaluation results for \ac{plmarker} fine-tuning are shown in Table \ref{tab:extraction_results}. Based on these results, we select the SciBERT-based \ac{plmarker} models to extract entities and relations from our corpus of \ac{nlp}-related research articles, resulting in large sets of entities and relations. To resolve duplicate entities, we use a rule-based approach that recognizes synonyms by unifying special characters and extracting abbreviations of terms that appear in parentheses immediately following an entity. In order to limit the set of eligible entities and relationships to high-quality ones, we select only those that are extracted more frequently than the thresholds of ${t}_{entities}=100$ and ${t}_{relations}=3$.

\paragraph{Manual Correction \& Construction} The extracted entities and relationships are passed to domain experts for validation and correction. In this case, the authors of the present work act as domain experts. If the domain experts consider a candidate triplet valid, it is manually inserted into the \ac{fos} hierarchy graph at the correct position. Otherwise, the candidate triplet is corrected, if possible, and only then inserted. Some candidate triples cannot be corrected since they involve out-of-domain terms, e.g., from the legal or medical field, and are, therefore, intentionally disregarded. 
Finally, we use GPT-4 \cite{openai2023gpt4} to generate short textual descriptions for each \ac{fos}. Table \ref{tab:fos_hierarchy_statistics} shows an overview of the resulting \ac{fos} hierarchy graph.

\begin{table}[!ht]
    \centering
    \resizebox{0.8\columnwidth}{!}{%
    \begin{tabular}{c|c|c}
    \toprule
       \textbf{\# Fields of Study} & \textbf{\# Relations}  & \textbf{Max Depth} \\
       \hline
        421 & 530 & 7 Levels\\
    \bottomrule
    \end{tabular}}
    \caption{Overview of the resulting \ac{fos} hierarchy graph.}
    \label{tab:fos_hierarchy_statistics}
\end{table}

\subsection{Fields of Study Classification}\label{sec:fos_clf}

To automatically assign research publications to the corresponding \acp{fos} in the hierarchy graph, we use a two-step classification approach. In the first step, we use the fine-tuned classification model of \citet{schopf-etal-2023-exploring}. It achieves an $F_{1}$ score of 93.21, using the 82 high-level \acp{fos} of the NLP taxonomy as classes, which we use as the starting point for our hierarchy graph. 

In the second step, we use the remaining \acp{fos} of our hierarchy graph as classes. Since we do not have sufficient annotated data to train a well-performing classifier, we use a rule-based approach. Thereby, publications are assigned to \acp{fos} depending on whether the stemmed \ac{fos} names or their stemmed synonyms are contained in the stemmed publication titles.

\subsection{Survey Paper Classification}\label{sec:survey_paper_clf}

To enable filtering by survey papers, we train a binary classifier that can automatically classify research publications into surveys and non-surveys. To this end, we construct a new dataset of survey and non-survey publications in \ac{nlp}. We obtain a list of candidate survey publications from keyword-based searches in the ACL Anthology and the arXiv cs.CL category using search terms such as \textit{"survey"}, \textit{"a review"}, or \textit{"landscape"}. We then manually annotate the candidate publications as positives if we consider them to be surveys based on their titles and abstracts. For negative sampling, we use the corpus of \ac{nlp}-related publications described in §\ref{sec:fos-construction}, excluding the previously identified positive examples. From this corpus, we randomly sample 15 times the number of positives as negatives to account for the inherent under-representation of surveys in conferences and journals. This annotation process results in a dataset of 787 survey and 11,805 non-survey publications in \ac{nlp}.

Using this survey dataset, we fine-tune and evaluate BERT, SciBERT, SPECTER2, and SciNCL models for binary classification. We create three different stratified 80/20 train/test splits and train all models for two epochs. Following the evaluation results in Table \ref{tab:survey_clf_results}, we select the SciNCL-based model as our final classifier.

\begin{table}[!ht]
    \centering
    \resizebox{\columnwidth}{!}{%
   

    \begin{tabular}{lc ccc ccc}
    \toprule
    
    \multicolumn{2}{l}{\textbf{Model $\downarrow$}} &      \textbf{Precision} &      \textbf{Recall} &    $\textbf{F}_{1}$ &      \textbf{Accuracy}  \\
    
    \midrule

    \multicolumn{2}{l}{BERT} &
    \textbf{84.35\small{$\pm$3.45}} &  77.49\small{$\pm$5.92} & 80.60\small{$\pm$2.07} & 97.68\small{$\pm$0.15}  \\
    \multicolumn{2}{l}{SciBERT} & 
    83.32\small{$\pm$2.21} &  82.38\small{$\pm$1.84} & 82.82\small{$\pm$0.81} & 97.87\small{$\pm$0.12} \\
    \multicolumn{2}{l}{SPECTER2} & 
    82.13\small{$\pm$4.58} & 85.77\small{$\pm$5.34} & 83.72\small{$\pm$0.38} & 97.92\small{$\pm$0.08}  \\
    \multicolumn{2}{l}{SciNCL} & 
    82.38\small{$\pm$4.01} & \textbf{86.53\small{$\pm$1.74}} & \textbf{84.35\small{$\pm$1.67}} & \textbf{98.04\small{$\pm$0.22}}  \\
  
    \bottomrule
    \end{tabular}
}
\caption{Evaluation results for survey paper classification as means and standard deviations on three runs over different random train/test splits. Since the distribution of classes is very unbalanced, we report micro scores.}
\label{tab:survey_clf_results}
\end{table}

\subsection{Additional Metadata}

To construct the \ac{nlp-kg}, we additionally use metadata obtained from the Semantic Scholar API. This includes short one-sentence summaries of publications (TLDRs), SPECTER2 embeddings of publications, author information, as well as citations and references. Further, we use PaperMage \cite{lo-etal-2023-papermage} to obtain the full texts of open-access publications.

\subsection{Semantic Search}

For semantic search, we use a hybrid approach that combines sparse and dense text representations to find the top-$k$ most relevant publications for a query. To this end, the results of BM25 \cite{10.5555/188490.188561} and SPECTER2 embedding-based retrieval are merged using Reciprocal Rank Fusion (RRF) \cite{10.1145/1571941.1572114}. 
To give more weight to the embedding-based approach, we set the $\alpha$ parameter determining the weight between sparse and dense retrieval to $0.8$. In addition, we use the S2Ranker \cite{Feldman_2020} to rerank the top $k=2000$ retrieved publications using additional metadata from the \ac{nlp-kg}, such as the number of citations and the publication date.


\subsection{Conversational Search} \label{sec:conversational-search}

To answer \ac{nlp}-related user questions and recommend relevant literature, we use the \ac{llm} in a \ac{rag} pipeline. Upon receiving a new user query, the \ac{llm} generates search terms using both the query and a one-shot example. These terms are then used for retrieving relevant publications via the semantic search module. Subsequently, the full texts of the top five search results are fed back to the \ac{llm}, which generates a response grounded in the retrieved literature. To make the generated answer verifiable for users and denote the knowledge sources, the \ac{llm} also generates inline citations. For follow-up queries, the \ac{llm} autonomously determines whether to respond using already retrieved publications or to initiate a new search. To reduce the hardware requirements of our server, we use the GPT-4 API for the conversational search and the \textit{Ask This Paper} feature.


\subsection{Ask This Paper}

In addition to the conversational search, the \ac{llm} integration enables user inquiries on specific publications via a popup window on each publication page. Users can either pose their own questions or choose from three predefined ones. Using the full text of the publication, the \ac{llm} generates verifiable answers supplemented by supporting statements, including section and page references from the publication text. Subsequently, the \ac{llm} generates three unique follow-up questions based on the conversation history.

\section{Demonstration}

\begin{figure}[h!]
    \centering
    \resizebox{\columnwidth}{!}{%
    \tcbincludegraphics[hbox, size=tight, graphics options={width=3cm}, colframe=black]{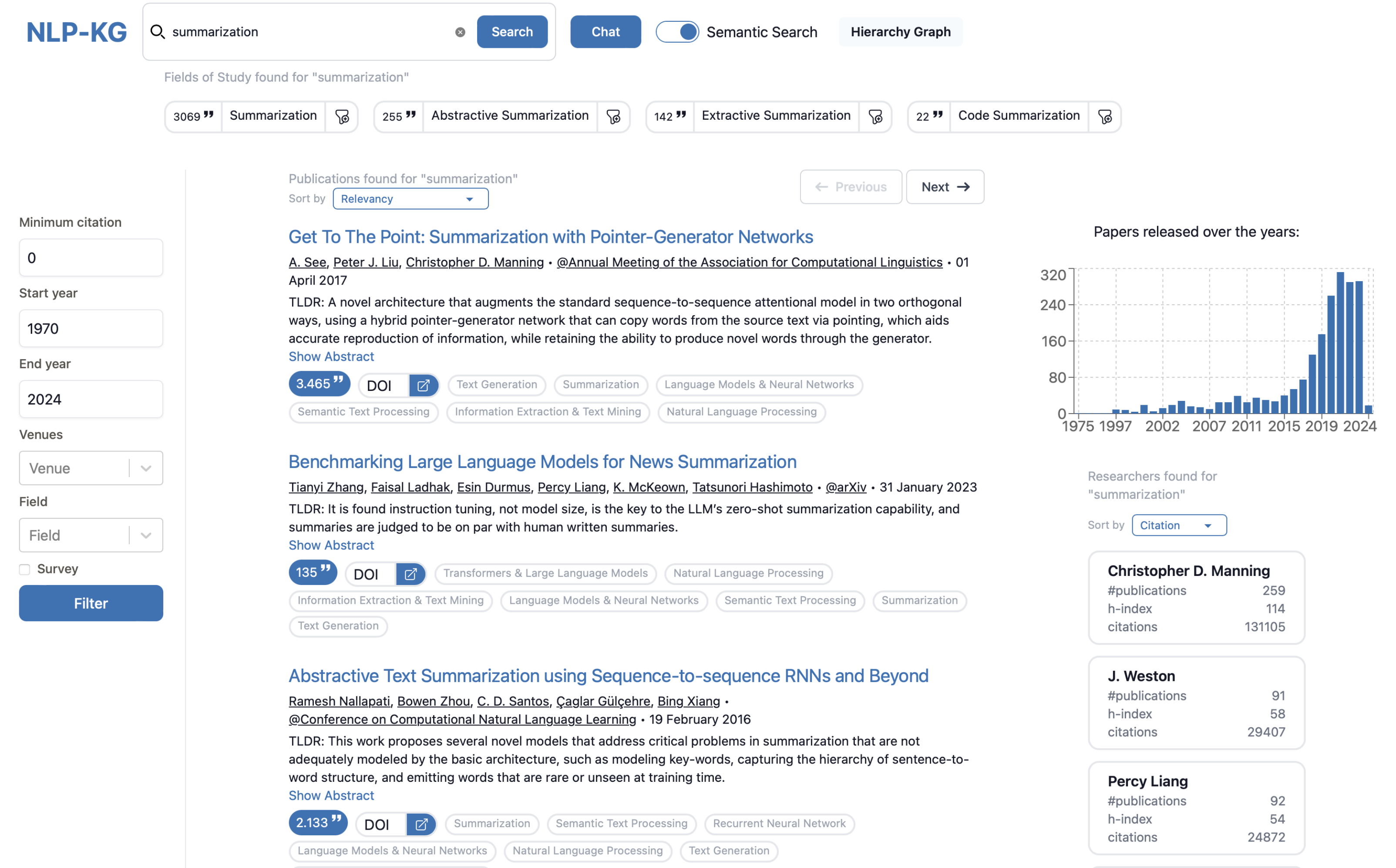}}
    \caption{Screenshot showing the semantic search and filtering features.}
    \label{fig:semantic-search}
\end{figure}

Our web application is built with Next.js\footnote{\href{https://nextjs.org}{https://nextjs.org}} and uses Python\footnote{\href{https://www.python.org}{https://www.python.org}} for the semantic search and preprocessing modules. The \ac{nlp-kg} is stored in Neo4j\footnote{\href{https://neo4j.com}{https://neo4j.com}} and the embeddings are stored in Weaviate\footnote{\href{https://weaviate.io}{https://weaviate.io}}. Our databases encompass publications from the entire ACL Anthology and the arXiv cs.CL category, enriched with metadata from Semantic Scholar. As illustrated in Figure \ref{fig:system-architecture}, the preprocessing module regularly fetches new publications, classifies them, and updates our databases.


Figure \ref{fig:semantic-search} shows the semantic search interface, allowing users to search for publications, authors, venues, and \acp{fos} using keywords via the top search bar. The central area shows retrieved publications, while relevant authors are listed on the right-hand side. Additionally, the top right corner showcases the annual publication count among the search results. On the left-hand side, users can access various filtering options, including the ability to filter by survey publications. Further, a list of \acp{fos} related to the search results is displayed at the top of the page, enabling users to navigate to dedicated \ac{fos} pages.


\begin{figure}[t!]
    \centering
    \resizebox{\columnwidth}{!}{%
    \tcbincludegraphics[hbox, size=tight, graphics options={width=3cm}, colframe=black]{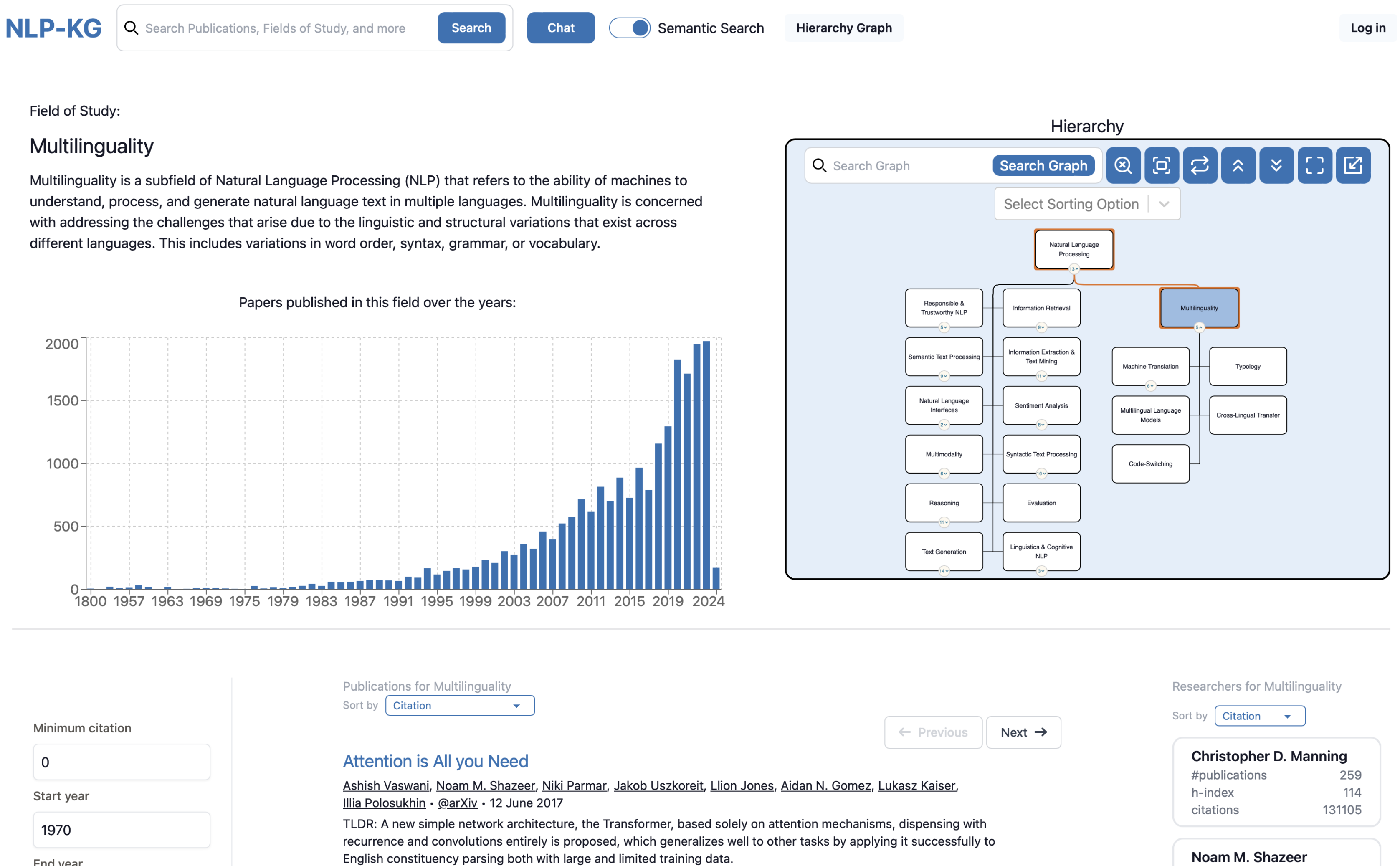}}
    \caption{Screenshot of the \ac{fos} view and the hierarchy graph visualization.}
    \label{fig:fos-page}
\end{figure}

\begin{figure}[b!]
    \centering
    \resizebox{\columnwidth}{!}{%
    \tcbincludegraphics[hbox, size=tight, graphics options={width=3cm}, colframe=black]{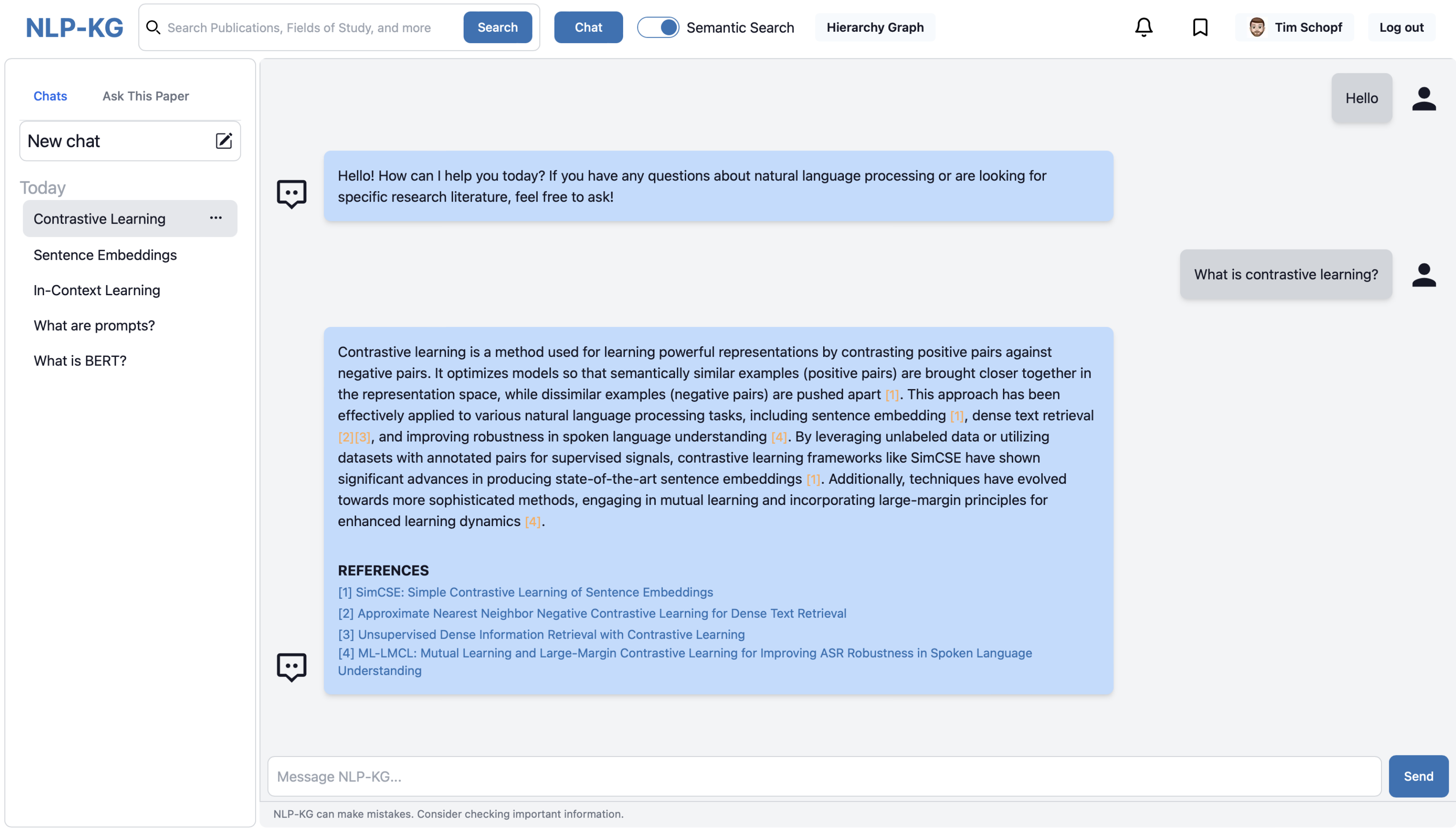}}
    \caption{Screenshot of the conversational search feature.}
    \label{fig:chat}
\end{figure}

Figure \ref{fig:fos-page} shows the \ac{fos} page, featuring a brief description of the respective \ac{fos} at the top, along with statistics on the annual publication count. The top right corner showcases a relevant section of the \ac{fos} hierarchy, enabling exploration of related fields. At the bottom of the page, users can explore and filter relevant authors and articles published on this topic.


Figure \ref{fig:chat} shows the conversational search feature. Users can pose \ac{nlp}-related questions to the \ac{llm}, which generates responses utilizing knowledge obtained from retrieved publications, accompanied by reference information. To enhance usability, the web application provides clickable links to referenced papers. Additionally, users can conveniently access their conversation history on the left-hand side.


\begin{figure}[h!]
    \centering
    \resizebox{\columnwidth}{!}{%
    \tcbincludegraphics[hbox, size=tight, graphics options={width=3cm}, colframe=black]{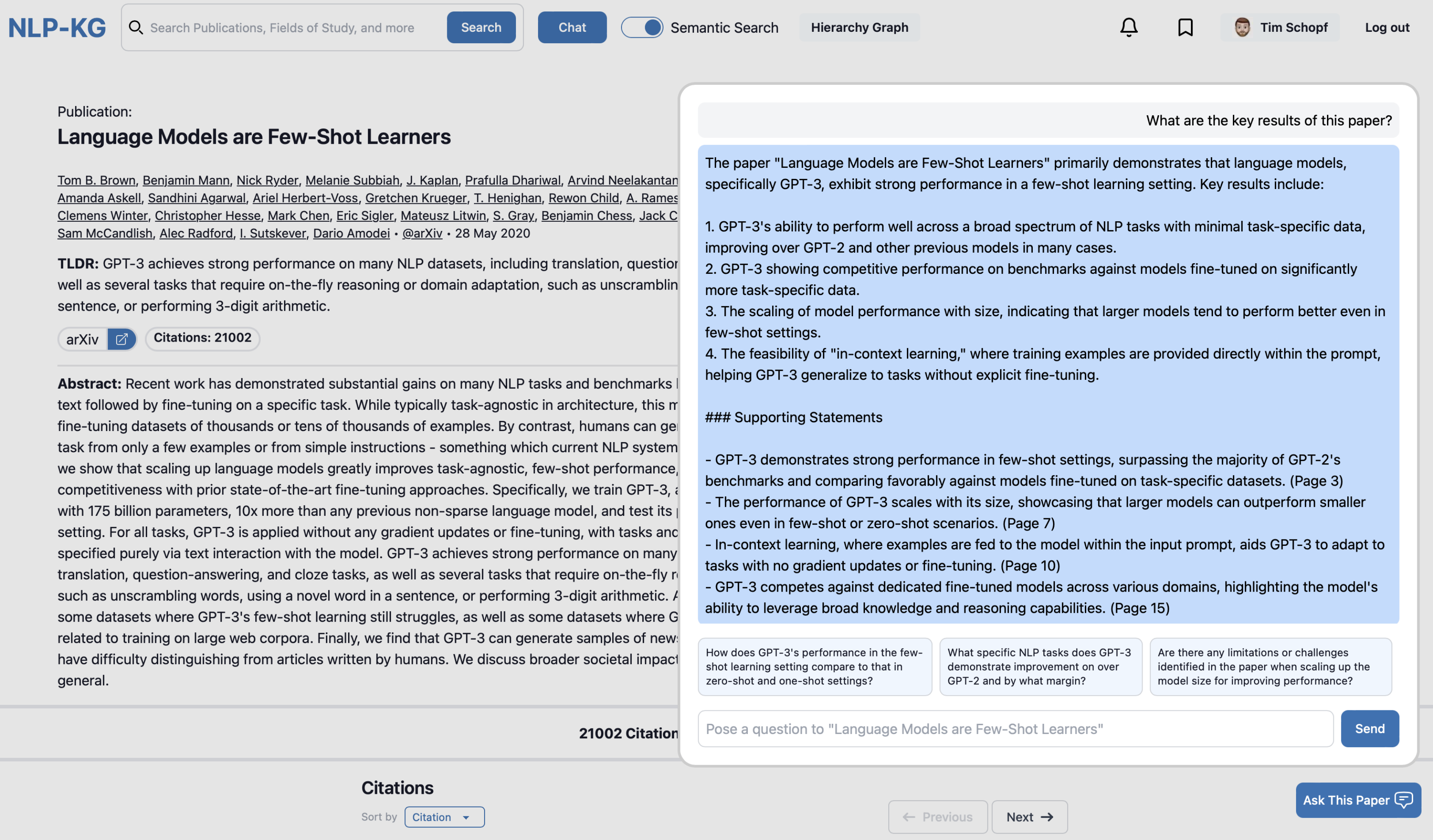}}
    \caption{Screenshot of the publication view and the \textit{Ask This Paper} feature.}
    \label{fig:ask-this-paper}
\end{figure}

Figure \ref{fig:ask-this-paper} shows the \textit{Ask This Paper} feature, enabling users to inquire about a specific publication. Accessible via a popup window at each publication page, users can choose from predefined questions or ask custom questions using the input field at the bottom of the chat window.


\section{Evaluation}

\subsection{Fields of Study Hierarchy Graph}

To evaluate the correctness of the \ac{fos} hierarchy graph, we conduct a user study involving ten \ac{nlp} researchers at the PhD level. Participants list five \ac{nlp} concepts related to their expertise while we ensure their presence in our graph. Subsequently, participants are presented with a visual representation of the constructed graph, initially showing only the first level of \ac{fos} in the hierarchy. This requires participants to expand the view by clicking to show the related \acp{fos}. Participants are then tasked with locating their provided \acp{fos} in the fewest steps possible, with each click or view extension counting as one step. Since the participants selected the \acp{fos} for the search themselves, we ensure their familiarity with the target field and related fields. We observe and count every step of the participants throughout their search process. Upon locating their \acp{fos}, participants evaluate the correctness of the relations utilized during their navigation and determine potential missing relations. Based on this assessment, we compute Precision, Recall, and $\textrm{F}_{1}$ scores, as shown in Table \ref{tab:relation_evaluation}, to evaluate the correctness of the traversed relations.


Furthermore, we use \ac{mape} to measure the percentage of errors or extra steps that participants make as they navigate the graph to reach their target \ac{fos}. We adopt the \ac{mape} metric as follows:

\begin{equation}
    \label{eqn:mape}
    \small
    \textrm{MAPE} = \frac{1}{n}\sum\biggl|\frac{\textrm{Total \#Steps - Ideal \#Steps}}{\textrm{Ideal \#Steps}}\biggr|,
\end{equation}

where $n=50$ denotes the number of \ac{fos} searches over all participants. In this context, a lower score means that, on average, users were able to find their target \ac{fos} with fewer extra steps. For example, a score of zero would mean that each user was able to find their target \ac{fos} with the optimal number of steps. Table \ref{tab:relation_evaluation} shows the evaluation results that demonstrate the high quality of the \acp{fos} hierarchy graph.

\begin{table}[!ht]
    \centering
    \resizebox{0.7\columnwidth}{!}{%
    \begin{tabular}{|c|c|c|c|}
    \hline
    \textbf{Precision} & \textbf{Recall} & $\textbf{F}_{1}$ & \textbf{MAPE} \\ \hline
    99.95 & 99.65  & 99.80 & 0.478 \\ \hline
    \end{tabular}}
\caption{Results for evaluating the correctness of relations in the \acp{fos} hierarchy graph.}
\label{tab:relation_evaluation}
\end{table}


\subsection{RAG Performance}

To evaluate the conversational search feature, we use the RAGAS framework \cite{es-etal-2024-ragas}, focusing on the \textit{Faithfulness} and the \textit{Answer Relevance} of generated responses. Faithfulness evaluates if the generated answer is grounded in the given context, which is important to avoid hallucinations. Answer relevance evaluates if the generated answer actually addresses the provided question. We use GPT-4 to generate 50 random questions related to \ac{nlp}, such as \textit{"Define perplexity in the context of language models"}. Subsequently, we utilize GPT-3.5 \cite{openai2022chat} and GPT-4 in our conversational search pipeline described in §\ref{sec:conversational-search} to generate grounded answers from retrieved publications. Finally, we use RAGAS to evaluate the generated responses. As shown in Table \ref{tab:rag-eval}, both \acp{llm} exhibit high faithfulness and answer relevance scores, indicating their ability to retrieve relevant publications from the \ac{rag} pipeline to effectively answer user queries based on provided contexts.


\begin{table}[!ht]
    \centering
    \resizebox{\columnwidth}{!}{%
    \begin{tabular}{lcc}
    \toprule
       \textbf{Model} & \textbf{Faithfulness} & \textbf{Answer Relevance}  \\
        \hline
        gpt-3.5-turbo-0125 & 0.9661 & 0.8479 \\
        gpt-4-0125-preview & 0.9714 & 0.8670 \\
    \bottomrule
    \end{tabular}}
    \caption{Evaluation results of our conversational search pipeline. Metrics are scaled between 0 and 1, whereby the higher the score, the better the performance.}
    \label{tab:rag-eval}
\end{table}

\subsection{Comparison of Scholarly Literature Search Systems}


We compare \ac{nlp-kg} with other publicly accessible systems for scholarly literature search, including Google Scholar, Semantic Scholar, \ac{orkg}, \ac{nlp} Explorer, and \ac{nlp} Scholar. A feature comparison is shown in Table \ref{tab:system-comparison}. 

\begin{table}[!ht]
    \centering
    \renewcommand{\arraystretch}{2} 
    \resizebox{\columnwidth}{!}{%
    \fontsize{18}{18}\selectfont 
        \begin{tabular}{|l|*{6}{>{\centering\arraybackslash}m{2.5cm}|}}
            \toprule
             & \textbf{Google Scholar} & \textbf{Semantic Scholar} & \textbf{ORKG} & \textbf{NLP \newline Explorer} & \textbf{NLP Scholar} & \textbf{NLP-KG} \\
            \hline
            Keyword-based Search & \cmark & \cmark & \cmark & \cmark & \cmark & \cmark \\
            \hline
            NLP specific & \xmark & \xmark & \xmark & \cmark & \cmark & \cmark \\
            \hline
            Fields of Study Tags & \xmark & \cmark & \cmark & \cmark & \xmark & \cmark \\
            \hline
            Fields of Study Hierarchy & \xmark & \xmark & \cmark & \xmark & \xmark & \cmark \\
            \hline
            Survey Filter & \cmark & \xmark & \xmark & \xmark & \xmark & \cmark \\
            \hline
            Ask This Paper & \xmark & \cmark & \xmark & \xmark & \xmark & \cmark \\
            \hline
            Conversational Search & \xmark & \xmark & \xmark & \xmark & \xmark & \cmark \\
            \bottomrule
        \end{tabular}%
    }
    \caption{Feature comparison of scholarly literature search systems.}
    \label{tab:system-comparison}
\end{table}

The comparison shows that \ac{nlp-kg} offers an extensive set of features providing users with a wide range of options to explore \ac{nlp} research literature. Unlike popular systems such as Google Scholar and Semantic Scholar, \ac{nlp-kg} is tailored specifically for \ac{nlp} research, ensuring an accurate and efficient exploration experience. Moreover, \ac{nlp-kg} is not limited to keyword-based searches, providing users with advanced search and retrieval features to explore the field of \ac{nlp}.

\section{Conclusion}

This paper introduces \ac{nlp-kg}, a system for search and exploration of \ac{nlp} research literature. \ac{nlp-kg} supports the exploration of unfamiliar fields by providing a high-quality knowledge graph of \acp{fos} in \ac{nlp} and advanced retrieval features such as semantic search and filtering for survey papers. In addition, a \ac{llm} integration allows users to ask questions about the content of specific papers and unfamiliar concepts in \ac{nlp} and provides answers based on knowledge found in scientific publications. Our model evaluations demonstrate strong classification and retrieval performances, making our system well-suited for literature exploration. 

\section*{Limitations}

The construction of the \ac{fos} hierarchy graph depends on the personal choices of the domain experts, which may bias the final result. The hierarchy graph may not cover all possible \acp{fos} and offers potential for discussions as domain experts have inherently different opinions. As a countermeasure, we automatically extracted entities and relations from a corpus of \ac{nlp}-specific documents and aligned the opinions of domain experts during the manual construction process.

We have limited the database of our system to papers published in the ACL Anthology and the arXiv cs.CL category. However, \ac{nlp} research is also presented at other conferences such as AAAI, NeurIPS, ICLR, or ICML, which may not be included in our system.

\section*{Ethical Considerations}

\ac{nlp-kg} supports the search and exploration of \ac{nlp} research literature in unfamiliar fields. To enable an intuitive user experience, the application integrates \ac{llm}-based features. However, \acp{llm} (e.g., GPT-4, used in this work) are computationally expensive and require significant compute resources. Additionally, although we aim to minimize model hallucinations by grounding the model responses in knowledge retrieved from scientific publications, the integrated \ac{llm} can nevertheless make mistakes. Therefore, users should always check important information provided by our \ac{llm}-based features. 

\section*{Acknowledgements}
Many thanks to Matthias Aßenmacher for his much appreciated proofreading efforts. We also thank Nektarios Machner, Phillip Schneider, Stephen Meisenbacher, Mahdi Dhaini, Juraj Vladika, Oliver Wardas, Anum Afzal, and Wessel Poelman for helpful discussions and valuable feedback. In addition, we thank Ferdy Hadiwijaya, Patrick Kufner, Ronald Ernst, Furkan Yakal, Berkay Demirtaş, and Cansu Doğanay for their contributions during the implementation of our system. Finally, we thank the anonymous reviewers for their useful comments.

\bibliography{anthology,custom}

\end{document}